\documentstyle[fullname]{article}

\newcommand{\eliza}{{\sc eliza}}
\newcommand{\parry}{{\sc parry}}
\newcommand{\Parry}{{\sc Parry}}

\makeatletter
\def\fnum@figure{{\bf Figure \thefigure}}
\makeatother

\title{Lessons from a\\Restricted Turing Test}
\author{Stuart M. Shieber \\
	Aiken Computation Laboratory \\
	Division of Applied Sciences \\
	Harvard University}
\date{March 28, 1994\\[.5ex](Revision 6)}

\begin{document}

\begin{titlepage}
\maketitle

\thispagestyle{empty}

\begin{abstract}
We report on the recent Loebner prize competition inspired by Turing's
test of intelligent behavior.  The presentation covers the structure
of the competition and the outcome of its first instantiation in an
actual event, and an analysis of the purpose, design, and
appropriateness of such a competition.  We argue that the competition
has no clear purpose, that its design prevents any useful outcome, and
that such a competition is inappropriate given the current level of
technology.  We then speculate as to suitable alternatives to the
Loebner prize.
\end{abstract}

\vfill

{\noindent This paper is to appear in {\it Communications of the
Association for Computing Machinery}, and is available from the Center
for Research in Computing Technology, Harvard University, as Technical
Report TR-19-92 and from the Computation and Language e-print server
as cmp-lg/9404002.}

\end{titlepage}

\section*{The Turing Test and the Loebner Prize}

The English logician and mathematician Alan Turing, in an attempt to
develop a working definition of intelligence free of the difficulties
and philosophical pitfalls of defining exactly what constitutes the
mental process of intelligent reasoning, devised a test, instead, of
intelligent behavior.  The idea, codified in his celebrated 1950 paper
``Computing Machinery and Intelligence'' \cite{turing-mind}, was
specified as an ``imitation game'' in which a judge attempts to
distinguish which of two agents is a human and which a computer
imitating human responses by engaging each in a wide-ranging
conversation of any topic and tenor.  Turing's reasoning was that,
presuming that intelligence was only practically determinable
behaviorally, then any agent that was indistinguishable in behavior
from an intelligent agent was, for all intents and purposes,
intelligent.  It is presumably uncontroversial that humans are
intelligent as evidenced by their conversational behavior.  Thus, any
agent that can be mistaken by virtue of its conversational behavior
with a human must be intelligent.  As Turing himself noted, this
syllogism argues that the criterion provides a sufficient, but not
necessary, condition for intelligent behavior.  The game has since
become known as the ``Turing test'', a term that has eclipsed even his
eponymous machine in Turing's terminological legacy.  Turing predicted
that by the year 2000, computers would be able to pass the Turing test
at a reasonably sophisticated level, in particular, that the average
interrogator would not be able to identify the computer correctly more
than 70 per cent of the time after a five minute conversation.

On November 8, 1991, an eclectic group including academics, business
people, press, and passers-by filled two floors of Boston's Computer
Museum for a tournament billed as the first actual administration of
the Turing test.  The tournament was the first attempt on the recently
constituted Loebner Prize established by New York theater equipment
manufacturer Dr.~Hugh Loebner and organized by Dr.~Robert Epstein,
President {\it Emeritus} of the Cambridge Center for Behavioral
Studies, a research center specializing in behaviorist psychology.
The Loebner Prize is administered by an illustrious committee headed
by Dr.~Daniel Dennett, Distinguished Professor of Arts and Sciences
and Director for Cognitive Studies, Tufts University, and including
Dr.~Epstein; Dr.~Harry Lewis, Gordon McKay Professor of Computer
Science, Harvard University; Dr.~H.  McIlvaine Parsons, Senior
Research Scientist, HumRRO; Dr.~Willard van Orman Quine, Edgar Pierce
Professor of Philosophy {\it Emeritus}, Harvard University; and
Dr.~Joseph Weizenbaum, Professor of Computer Science {\it Emeritus},
Massachusetts Institute of Technology.  (Dr.~I. Bernard Cohen, Victor
S. Thomas Professor of the History of Science {\it Emeritus}, Harvard
University, chaired the committee at an earlier stage in its genesis,
and Dr.~Allen Newell, U. A. and Helen Whitaker University Professor of
Computer Science, Carnegie-Mellon University, and the prize
establisher Dr.~Loebner served as advisors.)

The prize committee spent almost two years in planning the structure
of the tournament.  Because this was to be a real competition, rather
than a thought experiment, there would be several computer
contestants, and therefore several confederates would be needed as
well.\footnote{We follow the prize committee's terminology in using
the terms `confederate', `contestant', and `judge' for the computer
program entrants, the humans being compared against, and the human
interrogators performing the evaluation, respectively.  We use the
term `agent' for both confederates and contestants.} It was decided
that there would be ten agents all together.  In the event, six were
computer programs.  Ten judges would converse with the agents and
score them.  The judges and confederates were both selected from the
general public on the basis of a newspaper employment advertisement
that required little beyond typing ability, then screened by interview
with the prize committee.  They were chosen so as to have ``no special
expertise in computer science''.

The committee realized early on that given the current state of the art,
there was no chance that Turing's test, as originally  defined, had the
slightest chance of being passed by a computer program.  Consequently,
they attempted to adjust both the structure of the test and the
scoring mechanism, so as to allow the computers a fighting chance.  In
particular, the following two rules were added to dramatically
restrict Turing's test.

\begin{itemize}

\item	{\it Limiting the topic:} In order to limit the amount of area that
the contestant programs must be able to cope with, the topic of the
conversation was to be strictly limited, both for the contestants and the
confederates.  The judges were required to stay on the subject in
their conversations with the agents.

\item	{\it Limiting the tenor:} Further, only behavior evinced during the
course of a natural conversation on the single specified topic would
be required to be duplicated faithfully by the contestants.  The
operative rule precluded the use of ``trickery or guile.  Judges
should respond naturally, as they would in a conversation with another
person.''  (The method of choosing judges served as a further measure
against excessive judicial sophistication.)

\end{itemize}

\noindent As will be seen, these two rules --- limiting the topic and
tenor of the discussion --- were quite problematic.

The prize committee specified that there be independent referees
stationed in several locations: several in the rooms with the judges
and confederates to answer questions concerning interpretation of the
above rules, and one in the auditorium to serve as a sort of roving
ombudsman.  I was a referee in the confederates' room, and can vouch
for the fact that my and my colleagues' efforts there were hardly
needed; the confederates performed admirably.  Reports from the other
referees indicated the same for the judges.\footnote{The confederate
room referees, in addition to myself, were Susan Cole Dranoff, an
attorney at the firm of Ropes and Gray, and Dr.~Burton Dreben, Edgar
Pierce Professor of Philosophy {\it Emeritus}, Harvard University.
The judge room referees were Ned Block, Professor of Philosophy,
Massachusetts Institute of Technology, Robert {W.} Furlong, patent
attorney, and Dr.~Robert Harford, Professor of Radiology, Thomas
Jefferson University.  Dr.~Thomas Sheridan, Professor of Engineering,
MIT, served in the auditorium.}

Dr.~Loebner placed only two restrictions on the setting up of the
competition by the prize committee: that a competition be held each
year, and that a prize be awarded at each competition.  The prize at
this first competition was a nominal \$1500, although Dr.~Loebner has
reportedly earmarked \$100,000 for the first computer program to pass
the full Turing test at some later running of the competition.  (Costs
for the running of the competition itself were paid for by grants from
the National Science Foundation and the Sloan Foundation.)

\begin{figure}
\begin{center}
\begin{tabular}{rlllll|llll}
\multicolumn{10}{c}{\bf Rank Order of the Terminals}\\[2ex]
{\bf Least} & \multicolumn{8}{c}{\mbox{}}& {\bf Most}\\
{\bf Human-Like} & 1 & 2 & 3 & 4 & 5 & 6 & 7 & 8 & {\bf Human-Like}  \\
\cline{2-9}
&&&&&&&&& \\[-1ex]
& ${\cal B}$ & ${\cal A}$ & ${\cal E}$ & ${\cal D}$ & ${\cal C}$ & ${\cal F}$ &
${\cal H}$ & ${\cal G}$ \\
\end{tabular}
\end{center}
\caption{Mock-up of the form used to implement the scoring method for
the first Loebner competition. The judge writes the letters
corresponding to the terminals in order from least to most human-like,
and draws a line purporting to separate the computer contestants from
the human confederates.  In this case, the line has been drawn such that
three of the terminals (F, H, and G) were deemed to be connected to
humans.}
\label{fig:form}
\end{figure}

To determine the prize-winner, an ingenious scoring mechanism was
devised.  The Turing test involves a single binary decision, which is
either right or wrong.  But to determine a winner, the contestants had
to be ranked, so each judge was required to place all of the agents in
order from the apparently least human to most human.  This alone
induced the ordering on the basis of which the prize would be awarded.
The contestant with the highest average rank would be deemed the
winner of the tournament.  However, this does not allow a direct
reconstruction of the results of the 100 implicit binary decisions
that might be made: which of the agents were humans, and which
computers.  To allow for this to be deduced as well, each judge was
requested to place a single line separating the ranked agents into two
groups.  Those to the right of the line were claimed by that judge to
be humans, those to the left computers.  (See Figure~\ref{fig:form}.)
The judges were told that at least two of the agents were human
confederates, and at least two computer contestants, thus limiting the
number of places that the line could be (rationally) placed.  The
binary decisions could then be read off of the rankings by noting on
which side of the line each agent fell.  This demarcation process was
not used in the awarding of the prize, but was carried out for its
informational value alone.

\section*{The Event}

The tournament was to begin at 1 pm on the scheduled Friday.  One room
of the computer museum was set up with ten terminals for the judges,
each labeled with a code letter and the specified topic for conversing
with the associated agent.  In a back room, hidden from the publicly
accessible part of the museum for obvious reasons, five computers had
been set up to serve the four confederates.  (One terminal was
intended to be a backup, and in case it was not needed, to be
connected to a publicly accessible terminal so that press and the
public could interact with it as a sort of separate Turing test.)  In
a large auditorium, the ten conversations were projected each on its
own screen around the perimeter of the room, and A.~K.~Dewdney
provided running commentary.

Unfortunately, there were serious technical difficulties with the
rented computer equipment that had been set up for the confederates.
None of the three IBM computers could be made to appropriately
interact over the prepared lines with their companion terminal in the
judges' room.  (The two DEC workstations seemed to work fine.)  After
almost two hours of unsuccessful last minute engineering, the prize
committee decided to begin the competition with only two confederates
in place (just the number that the judges had been told was the
minimum), reducing the number of agents to eight.  The time that each
judge had to converse with each agent was shortened from approximately
fifteen minutes to approximately seven in order to accommodate the
press's deadlines.

The topics chosen by the six contestants were of the sort appropriate
for a cocktail party venue (burgundy wines, dry martinis, small talk,
whimsical conversation, dissatisfactions in relationships) or perhaps,
a child's birthday party (second grade school topics).  The two
participating confederates chose to converse on Shakespeare and women's
clothing.  In the end, and perhaps unsurprisingly, the average
rankings placed the two human confederates as ``more human-like'' than
the six contestants.  The highest-ranked contestant, Joseph Weintraub's
program (topic: whimsical conversation) was awarded the \$1500 prize.

The surprises, such as there were, were in the bipartite
classifications.  Five judges ranked the top contestant as human, and
there were eight instances of such misclassifications of computers as
humans overall.  (For the record, the contestant discussing relationship
problems, the entry of Dr.~Kenneth Colby, well known for his {\parry}
program, which simulates paranoid responses, was twice rated human.
The simulated seven-year old accounted for the final
misclassification.)

Ms.~Cynthia Clay, the Shakespeare aficionado, was thrice misclassified
as a computer.  At least one of the judges made her classification on
the premise that ``[no] human would have that amount of knowledge
about Shakespeare.''  \cite{wall-street} Ms.~Lisette Gozo was honored
as the most human of the agents for her discussion of women's
clothing, although one judge rated two computer programs above her.

Perhaps the most conspicuous characteristic of the six computer
programs was their poor performance.  It was widely recognized that
computer experts could readily distinguish the contestants from the
confederates.  Indeed, many of the techniques being used by the
programs were easily spotted by those familiar with the {\eliza}
program that prize committee member Weizenbaum developed in 1965.  The
repetition of previous statements verbatim (subject only to pronominal
adjustments, sometimes wrong), answers transparently keyed to trigger
words, and similar tricks of the {\eliza} trade were ubiquitous.  For
example, the following example from the whimsical conversation program
is illustrative of the regurgitation technique:\footnote{All excerpts
given below are taken verbatim from electronic transcripts of the
competition provided by and copyright 1991 of the Cambridge Center for
Behavioral Studies.  No changes were made except for the adjustment of
line breaks. In particular, spelling errors and extraneous characters
were let stand.}

\begin{quote}
{\it 01:11:42  Judge 10}\\
It has been a long day!  Now I think I'm ready for some fun.  The weather isn't
very pleasant today.  Do you have any jokes that could lift the clouds?\\[2ex]
{\it 01:13:01  Terminal 5}\\
 Everybody talks about the weather but nobody seems to do much
 about it. Why do you need to tell me now you think you're ready
 for some fun? Groannnnnn...............
\end{quote}

\noindent Note that the first sentence in the reply, which concerns
the weather, seems to be triggered by the keywords ``weather'' or
``clouds'' in the judge's contribution.  The second sentence is
comprised of the judge's second sentence, with pronouns transposed, and
preceded by the introductory ``Why do you need to tell me''.

Such techniques are subject to well-known and obvious errors, as can
be exemplified in the transcripts of interactions with the winning
computer entrant.  For instance, a naive application of pronoun
transposition interacts poorly with tag questions like ``aren't you''.
\begin{quote}
{\it 01:17:52  Judge 10}\\
You're trying to make this difficult for me aren't you?  I just explained that
I was ready for fun!\\[2ex]
{\it 01:18:53  Terminal 5}\\
Would you pre]fer it if it was not true that IO re trying to make
 this difficult for you aren't you?  I just explained that i was
 ready for fun! I am, therfore I compute.... What makes you think
 I am trying to make this difficult for you aren't i?
\end{quote}
Similarly, repetition of a compound phrase is conversationally bizarre.
\begin{quote}
{\it 01:35:26  Judge 6}\\
I can't say I understand the question...but the whimsy of a poet is
one thing and the whimsy of most of us is another.\\[2ex]
{\it 01:36:51  Terminal 5}\\
How do you know you can t say you understand the question...but
 the whimsy of a poet is one thing and the whimsy of most of us
 is another.
\end{quote}

One reporter noted that Weizenbaum himself was ``disturbed'' by how
easily people were fooled by these programs \cite{wall-street}, and
more than one of the judges reported that they were disappointed in
the programs' capabilities after their expectations had been raised by
interacting with {\eliza} in the interviewing process.  Dr.~Epstein,
in a speech after the event, noted that he had learned from the day's
proceedings that ``little progress has been made in the last
twenty-five years'', that is, since {\eliza}.  (We address this
dubious conclusion below.)

\section*{Analysis}

The obvious question, then, is how to reconcile the apparent success
of the programs in fooling judges with their patently low technology.
Clearly, part of the answer relies on the phenomenon that P. T. Barnum
used to amass a fortune.  People are easily fooled, and are especially
easily fooled into reading structure into chaos, reading meaning into
nonsense. This accounts for the popularity of newspaper horoscopes and
roadside psychics. This is not a flaw in the human mental capacity.
Sensitivity to subtle patterns in our environment is extremely
important to our ability to perceive, learn, and communicate.  Clouds
look like ships, and Rorschach blots seem like vignettes.  How much
different is interpreting non sequitur as whimsical conversation?

Ned Block, a professor of philosophy at MIT (and by coincidence a
referee at the competition, stationed with the judges) has argued that
the Turing test is a sorely inadequate test of intelligence because it
relies solely on the ability to fool people
\cite{block-computer-model}.\footnote{This is not the only case in
which exception has been taken to the appropriateness of the Turing
test as a barometer of intelligence.  See the discussion in
the next section.} Certainly, it has been known since
Weizenbaum's surprising experiences with {\eliza} that a test based on
fooling people is confoundingly simple to pass.

People are even more easily fooled when their ability to detect fooling
is explicitly vitiated, for instance, by a prohibition against using
``trickery or guile''.\footnote{Daniel Dennett, the head of the prize
committee, has himself argued against placing ``tacit restrictions on
the lines of questioning of the judges'', calling this a ``a common {\it
misapplication} of the sort of testing exhibited by the Turing test that
often leads to drastic overestimation of the powers of actually existing
computer systems.'' \cite[emphasis in original]{dennett-85}} When I
asked Mr.~Weintraub during the post-contest press conference how he
himself would have unmasked his program, his response --- typing
gibberish in to see if the program spat it back verbatim at a later time
a la {\eliza} --- was certainly outside the established rules.  In fact,
the referees had discussed that very technique the previous night at a
meeting with the prize committee to calibrate our collective
understanding of the rules.  I pointed out to Mr.~Weintraub that his
response fell under the ``trickery and guile'' prohibition, and he took
another stab at the question.  His second attempt to specify a winning
strategy against his program succumbed to the same problem.  (It
involved repeating questions multiple times.)

Weintraub's problem in answering the question points to the craftiness
of his solution to the Loebner prize puzzle.  His entry is unfalsifiable
independent of its performance and solely on the basis of the choice of
topic.  As almost everyone has noted who was familiar with the rules,
whimsical conversation is not in fact a {\em topic} but a {\em style} of
conversation (at least as practiced by Weintraub's program).  And
whimsical conversation in the mold of Weintraub's program is essentially
nonsense conversation, a series of non sequiturs.  Thus, when
Weintraub's program is unresponsive, fails to make any sense, or shows a
reckless abandonment of linguistic normalcy, it, unlike its competitor
programs, is operating {\em as advertised}.  It is being ``whimsical''.
At those times when, by happenstance, the program trips over an
especially suggestive response, a judge can grab at it as the real
article.  (The strategy is reminiscent of that used by the program
Racter to create ``free verse'' poetry, another unfalsifiable genre.)
Weintraub's strategy was an artful dodge of the competition rules.  He
had found a loophole and exploited it elegantly.  I for one believe
that, in so doing, he heartily deserved to win.

We might call this winning strategy ``{\parry}'s
finesse'',\footnote{Dennett \namecite{dennett-85} uses the term ``parrying''
for the Eliza-like technique of randomly generating a canned response
as an option of last resort, a key tool for implementers of {\parry's}
finesse.} after Kenneth Colby's previously mentioned {\parry} program
\cite{colby-81}.  {\Parry} was designed to engage in a dialogue in the
role of a paranoid patient.  The program was perhaps the first to be
subject to an actual controlled experiment modeled on the Turing test
\cite{colby-72}, in which psychiatrists were given transcripts of
electronically mediated dialogues with {\parry} and with actual
paranoids and were asked to pick out the simulated patient from the
real.  The fact that the expert judges, the psychiatrists, did no
better than chance, has been credited to the fact that
unresponsiveness and non sequitur are typical behaviors of paranoids.
Joseph Weizenbaum's response to the experiment --- in the form of his
own model of a deviant mentality --- parodies {\parry}'s finesse
succinctly:\footnote{Dennett \namecite{dennett-85} discusses this and other
problems with the {\parry} tests.  Arbib \namecite{arbib-74} presents a
contravening view, rejoined by
Weizenbaum \namecite{weizenbaum-74-reply}.}

\begin{quote}
The contribution here reported should lead to a full understanding of
one of man's most troublesome disorders: infantile autism\ldots.  It
responds {\it exactly} as does an autistic patient --- that is, not at
all\ldots.  This program has the advantage that it can be implemented on
a plain typewriter not connected to a computer at all.
\cite{weizenbaum-74}
\end{quote}

Post hoc thinking of this sort can go a long way to rationalizing the
various misclassifications of the whimsical conversation program or, in
the same vein, the program that talks at the level of a second-grader.
(Who could fail to give a seven-year-old child the benefit of the
doubt?)  It leads to noting other insidious forms of scoring bias that
crept into the competition.  One possible source of such bias, for
instance, follows from the technical problems that caused two of the
confederates to be eliminated.  Once the number of confederates had been
reduced to the announced minimum, it became impossible for a judge to
rationally place the demarcation line between ``humans'' and
``computers'' in such a way as to rate a human as a computer without
also rating a computer as a human.  Of course, the converse was not
true.  This might have accounted for one or two more of the errors.
Dr.~Epstein points out in response to this observation that ``(1) Two of
the ten judges drew the line after just {\em one} entry, in spite of our
instructions.  (2) Three of the 5 judges who mistook Weintraub's program
for a person rated it above one or both confederates.  (3) Two judges
mistook a confederate for a computer.  In fact, in two (and only two)
cases could our instructions have forced the judge to mistake a computer
for a person.'' (personal communication to Harry Lewis, 1992) The third
point is, of course, irrelevant, the first hardly gratifying, the second
accounted for by Weintraub's use of {\parry}'s finesse, and the final
comment is exactly my point.

But post hoc rationalization, like telling your boss off, may be
enjoyable at the moment, but is, in the long run, ungratifying.  The
important questions do not involve microanalysis of the particular
competition as run several months ago, but the larger questions of the
purpose, design, and even existence of the Loebner prize itself.

\section*{Why a Loebner Prize?}
\label{sec:why}

There is a long history of argumentation in the philosophical
literature opposing the appropriateness of the Turing test as a litmus
test of intelligence.  Certain arguments against the effectiveness of
the test in answering questions about the intelligence of computers or
the possibility of human thought center around the behaviorist nature
of the test.  Intelligence, it may be claimed, is not determinable
simply by surface behavior.  Variants of this argument have been given
by Block \namecite{block-psych}, Gunderson \namecite{gunderson}, and Searle
\shortcite{searle-chinese,searle-84}.   Others have suggested that the
Turing test is not sufficient in that the behaviors under adjudication
are too limited \cite{gunderson,fodor}.  On the basis of such
counterarguments, Moor \namecite{moor-analysis} has argued for a
drastically limited view of the Turing test, not as an operational
definition of intelligence at all, but rather as a mode for
accumulating evidence leading to an inductive argument for the
intelligence of the machine.  (See the reply by Stalker \namecite{stalker} and
a later clarification by Moor \cite{moor-explain} for further arguments.)
Moor \namecite{moor-encyc} provides a good introduction to these issues.
French \shortcite{french} provides a strong argument that as a
sufficient condition for intelligence, the Turing test is so difficult
as to be uninteresting.  Nonetheless, none of these sorts of
presumptive counterarguments to the use of a Turing test are the basis
for the discussion in the remainder of this paper.  The issue of
whether an operational definition of intelligence is appropriate, and
whether the particular definition codified in the Turing test is too
narrow, though important questions, can be taken as resolved in favor
of the Turing test for the purposes of the present discussion.  Thus,
we will side with the behaviorist interpretation favored by the
organization administering the prize, the Cambridge Center for
Behavioral Studies.  Nonetheless, these arguments do provide another
strong basis on which to question the appropriateness of the Loebner
prize.  A full discussion is, unfortunately, well beyond the scope of
this paper, but readers are urged to consult the cited literature.
Having sided, for the nonce, with the philosophical appropriateness of
Turing's design as a test of intelligent behavior, we turn to the
question of whether the Loebner prize competition is itself an
appropriate enterprise.

Prizes for technological advances have existed before, and much can be
learned by comparison with previous exemplars.\footnote{In fact, other
limited Turing tests have been carried out as well.  See the
discussion by Moor \cite[page 1129-30]{moor-encyc} for some examples.}
Just as humankind has dreamed of mimicking the human power of thought,
so have we longed to possess the avian power of flight.  Human-powered
flight entered the mythology of the ancient Chinese and Romans, the
designs of da~Vinci, yet was only accomplished within the last
generation as a direct result of a prize set up for the express
purpose of promoting that technology.  The Kremer prize, established
in 1959 by British engineer and industrialist Henry Kremer, provided
for an award of \pounds 5000 for the first human-powered vehicle to
fly a specified half-mile figure-eight course.  It was awarded in
1977, less than twenty years later, to a team headed by Paul Macready,
Jr., for a flight by Bryan Allen in the {\it Gossamer Condor}.

The success of the Kremer prize depended on two factors.

\begin{itemize}

\item	{\it Pursuing a purpose:} The goals of the Kremer prize were
clear.  At the time of the institution of the prize, there were no
active efforts to build human-powered aircraft.  The goal of the prize
was to provide an incentive to enter the field of human-powered
flight.  It was tremendously successful at this goal.  By the time
that the {\it Gossamer Condor} made its award-winning flight,
Macready's team was in competition with several other teams with
planes that were flying substantial distances solely under human
power.

\item	{\it Pushing the envelope:} The basic sciences underlying
human-powered flight were, by 1959, well understood.  These included
aerodynamics, mechanics, anatomy and physiology, and materials
technology.  It was even possible for Robert Graham, an expert in the
field of human-powered flight and a founding member of the Cranfield
Man-Powered Aircraft Committee, to state at that time that ``Man could
fly, if only someone would put up a prize for it.''  (Quoted by
Grosser \shortcite[page 23]{grosser}.)  Overcoming the human
difficulties in building a team that had collective mastery of these
various fields and the engineering difficulties in creatively
combining them were astonishing accomplishments.  Nonetheless, as it
turned out, no new basic discoveries were required at the time of the
founding of the Kremer prize to win it.\footnote{``The flight [of the
Gossamer Condor] has shown that, with what appears to be a
comparatively unsophisticated design, controlled man-powered flight
over a reasonable distance is possible.'' \cite[page 341]{reay}} The
task was just beyond the edge of the current technology.
Unfortunately, since our ability to dream far outstrips our ability to
build, the establishment of tests of ridiculous difficulty is not
difficult to imagine.  At a time when an award-winning human-powered
flight was one of one meter at an altitude of 10 centimeters (the 1912
{\it Prix Peugeot}), the Paris newspaper {\it La Justice} established
a prize for the first nonstop human-powered flight from Paris to
Versailles and back.  (It was never won.)

\end{itemize}

The history of human-powered flight indicates that only when the
purpose of the prize is clear and the task is just beyond the edge of
current technology is a prize an appropriate incentive.  The Kremer
prize is a prime example of a prize that meets these criteria.  The
Loebner prize is not.\footnote{Several other factors markedly
differentiate the Kremer and Loebner prizes.  First, whereas the
committee administering the Kremer prize consisted primarily of
scientists specializing in the engineering of human-powered aircraft,
it has been observed that current researchers in artificial
intelligence, computational linguistics, and natural-language
processing are conspicuous by their absence from the Loebner prize
committee.  (This problem has since been corrected.)  Second,
competition for the Kremer prize was on an as-needed, as opposed to
regular, basis, and no prize was awarded until the prize test was
completed in the presence of a qualified judge certified by the prize
committee.  Finally, the successful participants in the human-powered
flight competitions were uniformly groups with strong backgrounds in
the component technologies.  In the case of the Loebner prize, the
participants were almost without exception amateurs.}

We turn first to the goals of the Loebner prize.  It was, according to
the formal statement in the competition application,
``established\ldots to further the scientific understanding of complex
human behavior.''  Along these lines Dr.~Loebner has been quoted as
saying ``People had been discussing the Turing test; people had been
discussing AI, but nobody was doing anything about it.''
\cite{computerworld} The several thousand members of the American
Association for Artificial Intelligence may be surprised to learn that
nobody is doing anything about it.

Others have argued that the prize will serve to publicize the Turing
test, thereby increasing the public's awareness and understanding of
artificial intelligence.  Increased public understanding of AI is
certainly a laudable goal, especially since the regular appearance of
superficial popularizations in the press serves more to mislead the
public by alternately raising and dashing expectations than to inform it
by cogent coverage of actual results.  A flurry of the standard stories
in the press like ``Computer fools half of human panel'' \cite{sjmerc}
and ``Test a breakthrough in artificial intelligence'' \cite{herald} was
certainly one of the side effects of the Loebner prize competition, but
perhaps not a laudable contribution.

Overselling of AI by the media (and, occasionally,
practitioners\footnote{Dreyfus \shortcite{dreyfus} provides pertinent
examples.}) has, in its brief history, been a repeated and persistent
problem, and the hubristic claims of the organizers of the Loebner
prize that they are ``confident that within 10 to 20 years a system
will pass this electronic litmus test'' \cite{guardian} perpetuates
the hyperbole.  Robert Epstein, in his recent article describing the
event, its genesis, and his speculations as to its importance,
constructs a standard claim of this sort:

\begin{quote}
Thinking computers will be a new race, a sentient companion to our own.
When a computer finally passes the Turing Test, will we have the right
to turn it off?  Who should get the prize money --- the programmer or
the computer?  Can we say that such a machine is ``self-aware''?
Should we give it the right to vote? Should it pay taxes?  If you
doubt the significance of these issues, consider the possibility that
someday soon {\it you will have to argue them with a computer.}
\cite[emphasis in original]{epstein}
\end{quote}

Not surprisingly, the winner of the Loebner prize has jumped on the
publicity bandwagon by taking out an advertisement pushing his program
as the ``first to pass the Turing Test''.\footnote{Dr.~Dennett has, on
behalf of the Loebner prize committee, demanded that the advertising
claim be discontinued, at peril of lawsuit, and Weintraub has
apparently complied.} Conversely, a prize whose execution convinces
fellow scientists mistakenly that little progress has been made in a
quarter century does little to promote the field.  In summary, there
is a difference between publicity and increased public understanding.
Events of this sort --- and the Loebner competition has been no
exception --- tend to generate the former rather than the latter.

Dennett has hinted at a completely different goal for the Loebner
prize.  ``It is useful to have the demonstration of the particular
foibles that human beings exhibit in 1991\ldots.  We won't learn much
about AI from the Loebner prize for a long time, but we will learn
some non-negligible things about social psychology, perhaps, in the
meantime.'' (Dennett, personal communication) For instance, the
competition might be justified ``as a proving ground for the
environmental conditions necessary to permit the Turing test to
someday occur.  In other words, the Loebner competition can tell us
what we need to know about how humans behave in computer mediated
interactions.'' (Dranoff, personal communication) This line of
teleology for the Loebner prize, that it serves not as a test of the
abilities of the computers but of the psychologies of the various
participants, has often been proposed informally.  Such a ``conspiracy
theory'' of the prize as a vast psychology experiment executed on
unwitting and unconsenting adults is as unlikely as it is disturbing.
Of course, there is already an extensive literature on how humans
behave in computer-mediated interactions, and the Loebner competition
is not likely to contribute to it; it was not designed or executed as
a controlled scientific experiment, nor was that its apparent
intention, despite the hopes of Dennett and Dranoff that firm
conclusions in psychology might be gleaned from it.

Thus, it is difficult to imagine a clear scientific goal that the
Loebner prize might satisfy.  Turing's test as originally defined, on
the other hand had a clear goal, to serve as a sufficient condition
for demonstrating that a human artifact exhibited intelligent
behavior.  Even this goal is lost in the Loebner prize competition.
By limiting the test, it no longer serves its original purpose (and
arguably no purpose at all), as Turing's syllogism
fails.\footnote{Robert Epstein has claimed that ``We have changed the
Turing test as Turing would have if he were alive.'' \cite{guardian}
But it seems likely that Turing would have appreciated that the
limitations imposed on the test by the Loebner committee invalidate it
as even a sufficient criterion for intelligent behavior, and would not
have sanctioned such gross modifications.  An anonymous reviewer notes
that ``none of the conditions assumed by Turing are redundant for a
meaningful test --- not the unlimited domain, not the unlimited time,
not the interactive nature of the test, not the interrogator's full
awareness that one of the respondents {\it is} a machine.''} It is
questionable whether the notion of a Turing test limited in the ways
specified by the Loebner prize committee is even a coherent one.  The
prize committee spent some time with the referees attempting to
explicate the notion of ``natural conversation without trickery or
guile''.  It was suggested that a criterion be used as to whether you
might say the utterance in conversation with a stranger seated next to
you on an airplane.  For instance, what might a competition judge
legitimately ask on the topic of Washington, DC?  Certainly, the
question ``Are there any zoos in Washington?'' is the kind of thing
you might ask a stranger when flying to the capital for the first
time, whereas ``Is Washington bigger than a breadbasket?''  is just as
certainly a trick question.  What about ``Is there much crime in
Washington?''  Undoubtedly acceptable.  ``Are there any dogs in
Washington?''  An odd question for an airplane conversation.  ``Are
there many dogs in Washington?''  Sounds better.  ``Are there many
marmosets in Washington?''  Odd.  ``Are there many marmosets in the
Washington zoo?''  Okay again.  The exegesis of such examples begins
to sound like arguments about angels and sharp objects.

Similar problems accrue to the notion of limiting the topic of
discourse.  Is the last question about Washington, DC or marmosets?
(One of the referees in fact thought that this and similar questions
should be ruled out as it was not strictly on the topic of the city
alone.)  How about ``Are the buildings in Washington very modern?''
Perhaps a question about architecture, as the following question
surely is: ``Do you know any examples of neo-Georgian architecture in
Washington?''  Are culinary topics ruled out, as in ``What foods is
our nation's capital best known for?''  Such issues are not idle in
the context of the Loebner competition.  Cynthia Clay, the Shakespeare
expert, was asked why Mario Cuomo has been referred to recently as
``Hamlet on the Hudson''.  The question caused much consternation
among the referees peering over Ms.~Clay's shoulder.  Her response
was ``His brooding'' after which she coolly changed the topic back to
Shakespeare.  Or had it ever left?

The reason that Turing chose natural language as the behavior
definitional of human intelligence is exactly its open-ended,
free-wheeling nature.  ``The question and answer method seems to be
suitable for introducing almost any of the fields of human endeavor that
we wish to include.''  \cite[page 435]{turing-mind}  In attempting to
limit the {\em task} of the contestants through limiting the {\em domain
alone}, the prize committee succeeded in doing neither.

The distinction between domain and task is crucial.  Finance is a
domain, but not a task; withdrawing money from a bank account is a task,
one that is achievable through both human and computer intermediaries
these days; taking dictation of a funds-transfer request is a task that
only humans can currently undertake with reliability.  Had Babbage
limited his differential analyzer to multiply only even numbers, the
design would have been no more successful.  This is a limitation of
domain that does not yield a concomitant limitation in task.

It is well understood in the field that natural-language systems must
be tested using a constrained task.  Currently, standard limited tasks
can be found in evaluation of natural-language database retrieval
systems (like withdrawing money from a bank account on the basis of a
natural-language request) and speech recognition systems (like
transcribing a spoken funds-transfer request).  The tasks, typically
undertaken with limited vocabulary, are easily quantifiable along
several dimensions (for example, technical notions of precision,
recall, overgeneration, perplexity) independently of the subjective
judgments of lay judges.  In addition, they can be adjusted to sit
just at the edge of technology (a topic we return to below) unlike the
Turing test itself.  The natural-language-processing research
community has used such tests for some time now, and there has been
increased interest in issues of evaluation of systems (primarily at
the behest of funding agencies) over the last few years; whole
conferences have been devoted to the subject (see, for instance, the
report by Neal and Walter \namecite{nl-eval}).\footnote{Although the
limitations and evaluation methods may be more sophisticated, the use
of such task-limited evaluations to guide scientific research may be
no more beneficial.  (See the next section.)}

In summary, the Loebner prize competition neither satisfies its own
avowed goals, nor the original goals of Alan Turing. In fact, it is hard
to imagine a scientific goal that establishment of the Loebner prize
provides a better route to than would be provided by other uses of
Dr.~Loebner's \$1500, his \$100,000 promissory note, and the \$80,000 in
ancillary grants from the National Science Foundation and the Sloan
Foundation.  (Nonscientific goals are much easier to imagine, of
course.)

Now to the second criterion for an appropriate technology prize, that
the task be just beyond the edge of technology.  Imagine that a prize
for human-powered flight were set up when the basic science of the
time was far too impoverished for such an enterprise, say, in da
Vinci's era.  The da~Vinci prize, we shall imagine, is constituted in
1492 and is to be awarded to the highest human-powered flight.  Like
the Loebner prize, a competition is held every year and a prize must
be awarded each time it is held.  The first da~Vinci competition is
won by a clever fellow with big springs on his shoes.  Since the next
competition is only one year away (no time to invent the airfoil), the
optimal strategy is universally observed by potential contestants to
involve building a bigger pair of springs.  Twenty-five years later,
the head of the prize committee announces that little progress has
been made in human-powered flight since the first round of the prize
as everyone is still manufacturing springs.\footnote{Hubert Dreyfus
\shortcite[page 100]{dreyfus} has made a similar analogy of climbing
trees to reach the moon.}

Of course, a lot of progress had been made in human-powered flight in
those twenty-five years.  Da~Vinci himself was studying human
physiology and anatomy and the flight of birds, and --- although his
own work directly on the topic of human-powered flight, ornithopter
design, was essentially meritless beyond its decorative qualities ---
the apparently tangential work was, in the long run, pertinent to the
technologies that would eventually enable the {\it Gossamer Condor} to
be constructed.  (See, e.g., Gibbs-Smith \shortcite{davinci}.)
However, over that period, and indeed at every point during the
following four centuries, the kind of progress that needed to get made
to solve the problem was not directly observable {\it at that time} in
{\it incremental} improvement in solutions to the problem, the kind of
improvement that might be observable in an annual contest.
Nonetheless, tremendous scientific progress was made between the
fifteenth and twentieth centuries.  The {\it Gossamer Condor} and the
digital computer are two outgrowths of this progress.

The field of artificial intelligence is in that kind of
state.\footnote{Prize committee member Weizenbaum places the state of
AI technology a bit later in his analogy with Newtonian physics
\shortcite[page 199]{weizenbaum}, Dreyfus a bit earlier in his analogy
with alchemy \shortcite{dreyfus-alchemy}.  Neither writer is, of
course, sanguine about the prospects for progress in the coming
centuries.} The AI problem, like the problem of human-powered flight
in the Renaissance, is only addressed directly and dismissed as
imminently solvable by those who underestimate its magnitude.
Progress on restricted tasks in limited domains is well documented in
the literature on applications of artificial intelligence.  But
progress on the underlying science that has been made in the last
twenty-five years, important though it is, is not of the type that
allows incremental advantage to be demonstrated on the big problem,
the full-blown Turing test, nor should this be seen as a failing of a
field addressing a problem of the scope and magnitude of human
intelligence.  (And like all scientific endeavors, a lot of time can
be spent on fruitless avenues of attack; {\eliza}, as a discipline for
natural-language processing, was such a fruitless avenue.  It was
quite fruitful in other areas, however, as cogently argued by
Weizenbaum himself.)  Indeed, one aspect of the progress made in
research on natural-language processing is the appreciation for its
complexity, which led to the dearth of entrants from the artificial
intelligence community --- the realization that time spent on winning
the Loebner prize is not time spent furthering the field.

Twenty-five years of progress in the fields associated with the Turing
test --- artificial intelligence, computational linguistics, and
natural-language processing --- cannot be summarized in a single
program, but is captured in the many small results, some of which,
some day, at an unpredictable time in the future, may lead to a
dramatic demonstration of apparently intelligent artificial behavior.
To expect more is hubris.  What is needed is not more work on solving
the Turing test, as promoted by Dr.~Loebner, but more work on the
basic research issues involved in understanding intelligent behavior.
The parlor games can be saved for later.

\section*{Alternatives to the Loebner Prize}
\label{sec:alternatives}

Given that the Loebner prize, as constituted, is at best a diversion
of effort and attention and at worst a disparagement of the scientific
community, what might a better alternative use of Dr.~Loebner's
largesse be?  The goal of furthering the scientific understanding of
complex human behavior is no less laudable now than it was before the
competition, but clearly, a direct assault on a valid test of
intelligent behavior is out of the question for a long time; even the
prize committee well appreciates that.  Thus, any award or prize based
on a behavioral test must use a limited task and domain, so that the
envelope of technology is pushed, not ignored.  The efforts of the
Loebner prize committee to design such a test have failed in that the
test that they developed rewards cheap tricks like parrying and
insertion of random typing errors.  This is an (indubitably
predictable) lesson of the 1991 Loebner prize competition.

This problem is a general one: Any behavioral test that is
sufficiently constrained for our current technology must so limit the
task and domain as to render the test scientifically uninteresting.
Adjusting the particulars of the Loebner competition rules will not
help.  By way of example, many years of effort have gone into the
design of the tests of natural-language-processing systems used at the
annual DARPA-sponsored Message Understanding Conferences.  The trend
among entrants over the last several conferences has been toward less
and less sophisticated natural-language-processing techniques,
concentrating instead on engineering tricks oriented to the exigencies
of the restricted task --- keyword-spotting, template-matching, and
similar methods.  In short, this is because such limited tests are
better addressed in the near term by engineering (building bigger
springs) than science (discovering the airfoil).  Behavioral tests of
intelligence are either too hard for a prize or too rewarding of
incidentals.

At this stage, objective behavioral tests must give way to subjective
evaluative ones.  A more appropriate way to reward novel, potentially
breakthrough-inducing efforts toward the eventual goal of mimicking
intelligent behavior would be to institute a prize for just such
efforts, on the model of the Nobel prizes, ACM's Turing award, and
similar subjectively determined awards.  Rather than awarding lifelong
achievement or past accomplishments, however, the prize could be
awarded for particular discoveries, regardless of field, that the
committee determined were of sufficient originality, import, and
technical merit and that were deemed contributory to Turing's goal
(even though they may provide no incremental edge in a current-day
Turing test).  To avoid rapt obeisance to AI conventional wisdom, the
awards committee would include eminent thinkers from a wide range of
related fields (much as the current Loebner prize committee does) but
to ensure technical fidelity, a nominating committee of researchers
from the pertinent technical fields should verify purported results
before passing them on for consideration.  In order to prevent
degrading of the imprimatur of the reconstructed Loebner prize, it
would be awarded on an occasional basis, only when a sufficiently
deserving new result, idea, or development presented itself.  I am not
ostentatious enough to provide examples that I believe would be
appropriate for such an award; I am sure that the reader can imagine
one or two.\footnote{It is interesting to compare the Loebner prize
with the Leibniz award for automatic theorem proving, endowed in 1983
by the Fredkin Foundation and administered by Carnegie-Mellon
University.  Like the Loebner prize, the Leibniz award offers
\$100,000 on the basis of an extremely difficult task; it is to be
conferred on the occasion of the first major new mathematical theorem
whose proof is found with essential contributions by automatic theorem
proving.  However, there are important differences.  Awarding of the
Leibniz prize is at the discretion of the Committee on Automatic
Theorem Proving of the American Mathematical Society; it is therefore
a subjective test, as it must be to decide issues such as the
suitability of the theorem that was proved.  In the interim, until the
Leibniz prize is awarded, intermediate awards are occasionally (not
annually) presented.  The Milestone and Current Awards are conferred,
respectively, for ``foundational work in automatic theorem proving''
and for ``ongoing research that shows promise'', again at the
recommendation of the committee.  The Current Award, as an award for
present developments rather than past achievement, is therefore
structured in much the same way as the present proposal.}

As the years elapsed, and the speculations of this Loebner prize
committee as documented in their past decisions began to prove
perspicacious, the Loebner prize might grow in stature to that of the
highly sought prizes of other scientific areas, and so provide a
tremendous motivation for innovative ideas in the quest for an
artificial intelligence.

\section*{Postscript}

The Second Annual Loebner Prize Competition was held at the Cambridge
Center for Behavioral Studies on December 15, 1992.  The number of
computer entrants had decreased from six to three, with Joseph
Weintraub's program, complete with the winning strategy from the
previous year's competition, taking first prize once again, this time
under the purported topic ``men vs.\ women''.  Bigger springs had
prevailed.

\section*{Acknowledgements}

The research in this paper was supported in part by grant IRI-9157996
from the National Science Foundation and by matching funds from Xerox
Corporation.

I am grateful to the many readers of earlier drafts of this paper:
Ned Block,
Noam Chomsky,
Jacques Cohen,
Daniel Dennett,
Susan Cole Dranoff,
Barbara Grosz,
Harry Lewis,
David Mumford,
Fernando Pereira,
Jeff Rosenschein,
David Yarowsky,
and two anonymous reviewers.
I have incorporated many of their thoughtful comments into the paper,
although the opinions presented here are my own, and should not be
taken as necessarily representative of the previous readers' views.

\bibliographystyle{fullname}

\end{document}